\documentclass{article}

\PassOptionsToPackage{numbers, compress}{natbib}

 \usepackage[preprint]{neurips_2026}

\usepackage[utf8]{inputenc} % allow utf-8 input
\usepackage[T1]{fontenc}    % use 8-bit T1 fonts
\usepackage{hyperref}       % hyperlinks
\usepackage{url}            % simple URL typesetting
\usepackage{booktabs}       % professional-quality tables
\usepackage{amsfonts}       % blackboard math symbols
\usepackage{nicefrac}       % compact symbols for 1/2, etc.
\usepackage{microtype}      % microtypography
\usepackage{xcolor}         % colors
\usepackage{graphicx}
\title{LEVANTE-bench: Multi-Scale Comparison of VLMs to Children Using Cognitive Tasks\\ (or, ``Is Your VLM Smarter Than a 5th Grader?'')} 
\author{%
  Alvin Wei Ming Tan \qquad David Cardinal \qquad Tania Lorido-Botr\'{a}n\\
  \textbf{Laura Bravo-S\'{a}nchez \qquad Sunny Yu \qquad Michael C. Frank}\\
  Stanford University\\
  \texttt{\{tanawm, david81, botran, lmbravo, syu03, mcfrank\}@stanford.edu}
}

\begin{document}

\maketitle

\begin{abstract}
Given the inherently multimodal nature of human experience, vision--language models (VLMs) hold substantial promise for modeling human cognition as it grows and develops with experience. Realizing their potential requires tools for comparing VLMs with human cognitive development across tasks, ages, and populations. We present LEVANTE-bench, a benchmark based on tasks and data from the Learning Variability Network (LEVANTE), which distributes open-source tasks and data measuring children's cognition across languages and cultures. In LEVANTE-bench, we systematically assess VLMs on six tasks, comparing their alignment with children aged 5--12 ($N$ = 1547) across three countries. We compare models at multiple scales, assessing their overall accuracy, their task- and item-level alignment with children, and how well they match children's trial-level error distributions. Alignment was heterogeneous across scales: at the level of tasks and items, more capable models aligned better with humans. However, match to human error distributions varied widely across tasks, and for several tasks, smaller models matched younger children's errors better. In addition, even the best-performing VLMs struggled on matrix reasoning and mental rotation tasks. Thus, current VLM architectures align only partially with the cognitive abilities of children. 
\end{abstract}

\section{Introduction}

Artificial intelligence models hold substantial promise as scientific tools for understanding human learning and cognition \citep{simon1980cognitive,rumelhart1986parallel}. Although the rise of language models has led to an explosion of cognitive evaluation work \citep{webb2023emergent,binz2023using,frank2025cognitive}, the unimodal nature of such models is an important limitation in using them to study and understand human cognition \citep{pavlick2023symbols}. In particular, for researchers interested in the efficiency and robustness of human learning, it is striking that language models are trained on much \emph{more} language than humans and much \emph{less} data of any other type \citep{frank2023bridging,warstadt2023findings}. Visual experience is an especially rich form of data that enables learners to acquire information about the causal structure of the world \citep{ayzenberg2025fast,huber2023developmental,aw2026zeroshotworldmodelsdevelopmentally}.

\begin{figure}
  \centering
  \includegraphics[width=\textwidth]{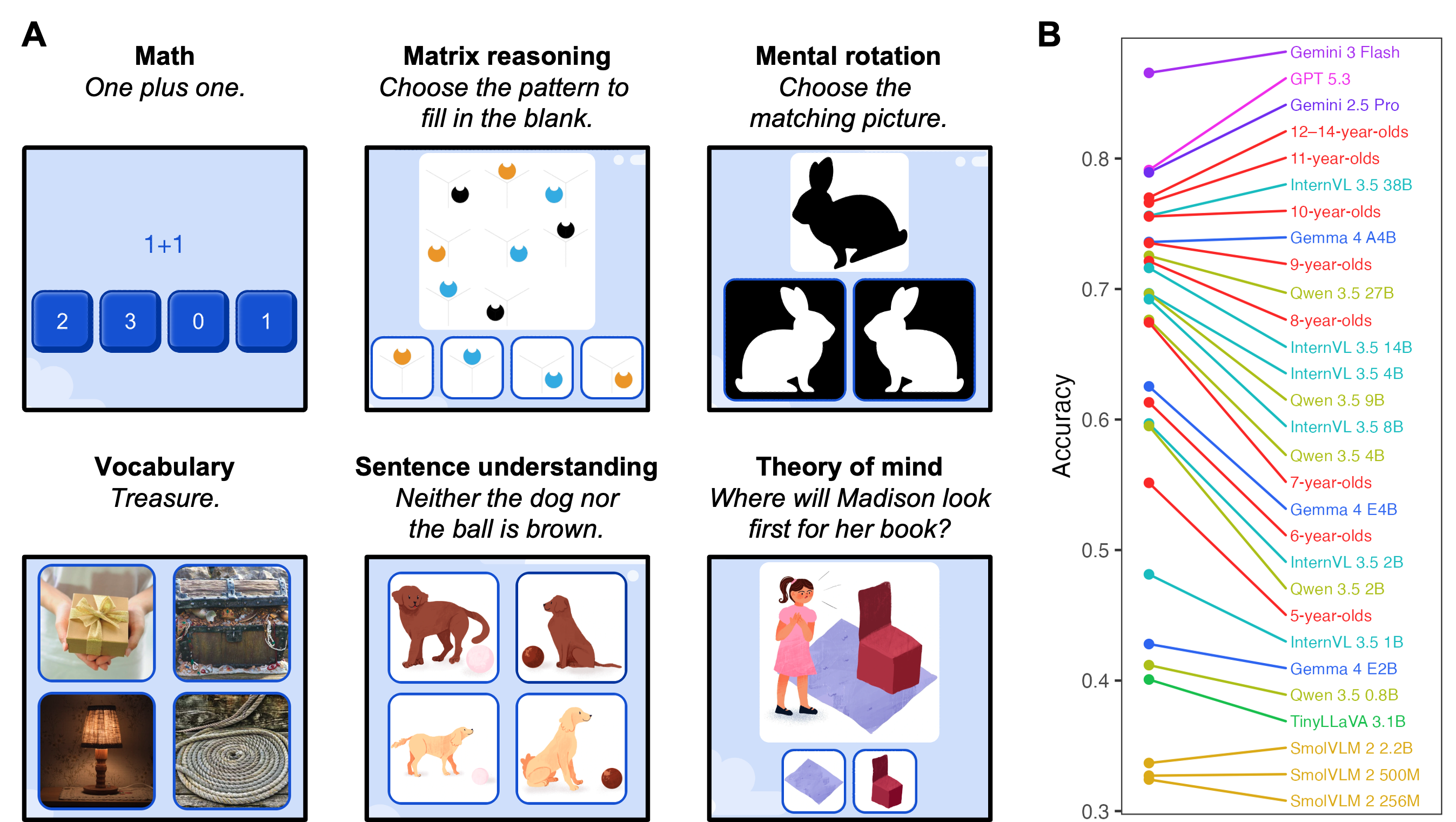}
  \caption{(A) Example items from the six tasks of LEVANTE-bench as presented to human participants; adapted from \citep{kachergis2025creation}. (B) Overall accuracies of models (colored by family) and humans (\textcolor{red}{red}).}
  \label{fig:graph_abs}
\end{figure}

Vision--language models (VLMs) \citep{radford2021learning, liu2023visual} thus offer important opportunities for cognitive modeling, especially for understanding human development. First, VLMs can be compared to human cognitive abilities \citep{schulze2025visual} and can even be pre-trained on human visual experiences \citep{vong2024, vong2025robustness, wang2025babyvlm, wang2025bbabyvlm}. In addition, VLMs can be evaluated using the multimodal format that is most common in experiments with children \citep{fernald2008looking,frank2023baby}. These two observations mean that VLMs can potentially be used to simulate important aspects of human development, allowing inferences about what behaviors arise from powerful statistical learning mechanisms and rich experiences. The promise of such models is that they can formalize and implement scientific theories from cognitive science, such as helping us better understand which aspects of cognitive development are innately specified \citep{vong2024, elman1996rethinking, singh2026innateness}. 

For VLMs to be used as cognitive models of learning, they must be evaluated in parallel with human behavioral performance -- ideally, on learning trajectories from children. Many such efforts leverage experiments on visual cognition in adults \citep{schulze2025visual,cao2025visualcognitiongaphumans, stogiannidis2025mindgapbenchmarkingspatial}. However, some recent developmentally-inspired VLM benchmarks have assessed models on concepts and domains typically studied in children  \citep{wang2025babyvlm,li2025coreknowledgedeficitsmultimodal, chen2026babyvisionvisualreasoninglanguage}, and some have compared VLMs directly to data from children \citep{wang2025babyvlm, tan2025assessing, yiu2025kivakidinspiredvisualanalogies}. 

However, comparisons are limited by the challenges of collecting child data. Most datasets are English-only, decreasing the generalizability of cognitive comparisons \citep{blasi2022over}. Few benchmarks compare humans and models at the item level; instead, most only assess overall accuracy \citep{tan2025devbench}. In addition, none of these benchmarks use tasks that have been psychometrically assessed for reliability and validity \citep{truong2025fantasticbugsaibenchmarks}. Finally, very few include data from multiple tasks on the same sample of children. This last point is especially important given that making cross-task comparisons may be relatively meaningless if the tasks are calibrated on different samples. The current paper aims to fill these gaps. 

We take advantage of a new resource: the Learning Variability Network Exchange (LEVANTE) \citep{frank2025learning}. LEVANTE provides a set of tasks for measuring children's learning and development, a task administration framework for researchers, and a global dataset of observations gathered from these tasks. The LEVANTE core tasks, in particular, are a set of psychometrically validated tasks that can be used to study learning development in children aged 5--12, spanning math, executive function, reading, language, spatial cognition, social cognition, and reasoning \citep{kachergis2025creation}. These tasks are hosted on a central platform, so that as researchers use them for data collection, their data flows into a centralized repository for open distribution. All LEVANTE data and task assets are licensed for non-commercial usage, meaning that they can be reused for VLM evaluation. Further, all tasks are typically given to the same set of children, meaning that task difficulty between children can be fairly compared.

In this work, we construct an evaluation benchmark (LEVANTE-bench) allowing systematic comparison of VLM performance with human cognitive development across tasks spanning math, reasoning, language, and social cognition (Figure~\ref{fig:graph_abs}A) and across three languages (English, Spanish, and German), using data from more than 1500 children. Tasks vary in difficulty: the easiest can be solved by small-scale open models, while the hardest are still challenging for current commercial frontier models. Our first contribution is thus to provide the largest and most comprehensive dataset to date for comparing VLMs to children's cognition.  

Our second contribution is to provide a framework for multi-scale comparison of model and humans. In particular, we measure model--human alignment across three scales: alignment on task difficulties, alignment on item difficulties within tasks, and alignment on trial-level error distributions.\footnote{We use the term ``alignment'' to denote a high level of correspondence between human and model on a particular metric, rather than to denote correspondence specifically on goals and values (as the term is used in discussions of AI safety).} The results of this analysis suggest that alignment is varied across scales: while larger models broadly align on task difficulty, all models are at best modestly aligned on item difficulty and are heterogeneous in their trial-level alignment. Together these results highlight gaps in VLM cognitive alignment.

\section{Prior work}

A vast literature compares text-only language models with human cognition; overall, human--model behavioral alignment for models is striking \citep{webb2023emergent,xie2025using,hewitt2024predicting} and improves with fine-tuning \citep{binz2023using}, though there are certainly still areas of lower alignment \citep{hu2025re, xie2026aipsychobench, mckenzie2023inverse}. A smaller but still extensive set of benchmarks and studies explicitly compare learning trajectories to human development \citep{shah2024development,yiu2024transmission}; many of these contain linguistic reframings of tasks that are typically given to children in multimodal formats. Theory of mind evaluation is an example of such an evaluation: Children's understanding of others' beliefs is typically assessed using picturebook tasks \citep{wellman2001meta}, but the vast majority of LLM theory of mind evaluations are in text-only formats \citep{hu2025re, kosinski2024evaluating}. 

Our goal here is to quantify the broad developmental alignment between VLMs and children in multiple domains of cognition. 
Most relevant to our current work are developmental comparisons of VLMs to human data and phenomena. Several of these are concentrated in specific domains, such as visual cognition \citep{li2025coreknowledgedeficitsmultimodal, chen2026babyvisionvisualreasoninglanguage, weihs2022benchmarking, sheybani2024modelvsbaby}, word learning \citep{jiang2023mewl}, language \citep{tan2025devbench}, and relational reasoning \citep{yiu2025kivakidinspiredvisualanalogies}. Most of these do not seek broad coverage over multiple domains, with the exception of \citet{wang2025babyvlm}, who compare VLM abilities to tasks described by the NIH Baby Toolbox, a broad coverage tool for measuring cognition in early childhood \citep{gershon2024nih}; however, data and stimulus accessibility issues limit their ability to directly compare children and models. Perhaps most closely related, \citet{tan2025devbench} proposed a method for comparing children's response distributions to those of models and applied it across a range of language tasks. Unlike LEVANTE-bench, these works lack human data spanning across ages and tasks, limiting the strength of cross-task comparisons that can be performed.
 
\section{Human tasks and data}
\label{sec:human}

LEVANTE includes an open collection of measures of child learning and development designed to be useful in multiple countries, cultures, and languages \citep{frank2025learning}. All tasks were designed for children aged 5--12 years with a planned downward extension from 2--5 years. Currently, the tasks are available in English, Spanish, and German. All data collected through the LEVANTE framework are released for open reuse on Redivis.\footnote{See \url{http://researcher.levante-network.org} for more details about the project, tasks, and data. Note that use of LEVANTE data requires affirmation of a data use agreement.} We selected six tasks from the LEVANTE core task battery \citep{kachergis2025creation} based on their suitability for processing by VLMs and their use of a simple multiple-choice format (see Figure~\ref{fig:graph_abs}A): 

\begin{itemize}
\item {\bf Math}. Each item is a math question that includes simple 4-alternative forced choice (4-AFC) number identification, comparison, arithmetic, and fraction problems. Number-line items were excluded because they did not use a multiple choice format.
\item {\bf Matrix reasoning}. Each item is a 3$\times$3 matrix of images that form a pattern, with the lower right element omitted. The participant must deduce which option in four fits the pattern (4-AFC).
\item {\bf Theory of mind}. Each item is a story with a 2-, 3- or 4-AFC question tapping reasoning about the beliefs and emotions of the individuals in the stories. 
\item {\bf Mental rotation}. Each item is a shape (either a 2D silhouette or a 3D shape) that must be rotated to match to one of two targets (one rotated and one mirrored; 2-AFC.
\item {\bf Sentence understanding}. Each item is a sentence that must be matched with one of four pictures (4-AFC).
\item {\bf Vocabulary}. Each item is a word that must be matched with one of four pictures (4-AFC).
\end{itemize}

We use human response data from the LEVANTE 2026.1 data release, which contains data collected from 5--12-year-old children in Colombia ($N$ = 1020), Canada ($N$ = 188), and Germany ($N$ = 339) for a total of 309,108 trial-level responses across all tasks.\footnote{The larger number of participants in Colombia is because not all participants performed all tasks in this context, thus more participants were collected to reach the target participant count in each task.} Data in Colombia were collected in schools; data in Canada were collected in a laboratory setting; and data in Germany were collected remotely at the participants' homes. See \citet{kachergis2025creation} for more details on data collection.

\section{Human--Model Comparison}

LEVANTE tasks contain items with a wide range of difficulty suitable for both younger and older children. To avoid boredom and frustration, not all participants saw all items; instead, after an initial period of normative data collection, tasks were made adaptive so that items were selected online based on their estimated difficulty for each task taker. Individual children's scores are thus not determined based on their proportion of correct answers on all items, but instead using an item-response theory (IRT) model that assigns an ability score $\theta$ determined by fitting a model of the form $P(r_{i,j}=1 \,\vert\, \theta_i, \delta_j) = \frac{e^{\theta_i - \delta_j}}{1 + e^{\theta_i - \delta_j}}$, where $r_{i,j}$ is the response for participant $i$ and item $j$ \citep{embretson2013item,truong2026roadmap}.\footnote{In practice, scoring models for the LEVANTE core tasks make use of either the Rasch model (as described in the text) or a 2-parameter logistic model (2PL) for the mental rotation task, and all models also included a per-item guessing lower bound which indicated the chance level for each item based on how many options were available. All models except models for the vocabulary task are multi-group IRT models selected based on model comparison \citep{bock1997multiple}; see \citep{kachergis2025creation} for details.} 

\paragraph{Cross-task alignment.} To estimate task difficulty, we used the $\theta$ and $\delta$ values to estimate human performance on all trials for all participants, even trials that they did not see. We then estimated the mean accuracy across all trials for each task, and correlated model and human task accuracies. 

\paragraph{Within-task alignment.} We compared human item difficulty $\delta$ to model accuracy, calculated as the proportion of runs in which models correctly responded to a given trial. For ease of interpretation, we negated item difficulty values to obtain item easiness values, such that greater model--human alignment would be indicated by higher correlations.

\paragraph{Trial-level error alignment.} We also make use of raw LEVANTE trial data to recover item-level empirical response distributions for each item over the available answer options. These data allow us to evaluate models based not just on their accuracy, but also on the similarity of their answer distributions with human response patterns. In particular, we use a method similar to \citet{tan2025devbench}: for each item, we computed the Kullback--Leibler divergence $D_\textup{\small KL} (d \parallel m)$ between the estimated multinomial response distributions for a model $m$ (empirically estimated from multiple runs) and human response distributions $d$, which we treat as the ground truth distribution. We added $\epsilon = 10^{-12}$ to the proportion of each option to smooth the distributions.
Because higher-ability children were more likely to see more difficult trials (e.g., older children are the only ones who see fractions trials in the math task), human response distributions are not comparable for items of different difficulty. To avoid this issue, we stratify response distributions by participant ability ($\theta$) and consider the similarity of the model to the response distributions of participants at the same general level.  

% For the six tasks, we evaluated our Redivis dataset contained  from the three LEVANTE pilot sites. <\textbf{Something about IRT models and item difficulty}>

\section{Experimental setup}

\paragraph{Models.} Our goal is to be able to compare models with humans within model families and across model scales. We thus selected a variety of open-weight VLM families that span a range of parameter sizes:
% possibly just turn this into an in-text list if tight on space
% \begin{itemize}
%     \item 
    Gemma 4 (E2B, E4B, 26B, 31B);
    % \item 
    InternVL 3.5 (1B, 2B, 4B, 8B, 14B, 38B);
    % \item 
    % Molmo 2 (4B, 7B, 8B);
    % \item 
    Qwen 3.5 (0.8B, 2B, 4B, 9B, 27B);
    % \item 
    SmolVLM 2 (256M, 500M, 2.2B);
    % \item 
    and TinyLLaVA (3.1B).
% \end{itemize}
Models were run on a combination of compute resources depending on availability, including an NVIDIA GeForce RTX 3090, several NVIDIA A40s, and an NVIDIA DGX H100 supercomputing cluster, or via the HuggingFace API. We also ran three closed-weight commercial frontier models via API to provide a topline comparison: GPT 5.3, Gemini 2.5 Pro, and Gemini 3 Flash.\footnote{For closed-weight models, we used estimates of their number of parameters from \cite{li2026incompressibleknowledgeprobesestimating, sturgeon2026}; we acknowledge that these estimates are preliminary and noisy, but they likely do not make a significant difference to our results because they are definitely at least an order of magnitude larger than the open-weight models we evaluated.}

\paragraph{Evaluation configuration.} To estimate error distributions and minimize response bias effects \citep{dominguez2024questioning}, we evaluated each model a minimum of 10 times on each item, permuting response options between runs. We ran all models with max tokens set to 1024 for open-weight models and 2048 for commercial models. We used thinking mode and FlashAttention-2 for models that support them, and 16-bit floating point precision for consistency. Images were resized to 512$\times$512 pixels. All experiment code as well as downloaders for LEVANTE assets and data can be found on \href{https://anonymous.4open.science/r/levante-bench-3013/}{GitHub}.

\paragraph{Question formatting.} Item prompts for the LEVANTE tasks were adjusted from the original tasks to account for differences in presentation format (e.g., non-sequential presentation). Prompt text (and the prompt image(s), if relevant) was passed to the model along with the possible response options. Response options were randomized and each option was labeled with a single capital letter (A through D). Most models were then instructed to provide a response in JSON format with the keys ``answer'' (a single capital letter) and ``reason'' (open-ended text); these responses were parsed with a parser that incorporated light error correction. Some of the smaller models (SmolVLM 2) were unable to accurately return JSON; instead, we instructed those models to return a single capital letter.

\paragraph{Prompt sensitivity study.} Prior to the main evaluation, we conducted an extensive task prompt sensitivity study spanning 5~model families (Qwen, InternVL, Gemma, SmolVLM 2, SpaceThinker) and all six tasks (Appendix~\ref{sec:prompt_sensitivity}).
We tested a wide range of strategies -- structured layouts, chain-of-thought (CoT), few-shot exemplars, self-consistency, task-specific expert framing, and elimination instructions -- and found three consistent patterns.
First, the optimal prompt varied by model family and scale: enriched prompts that helped larger models ($\geq$4B) often hurt smaller ones.
Second, CoT and other reasoning elicitation strategies frequently \emph{decreased} accuracy for sub-4B models substantially, while also lowering parse rates.
Third, a minimal prompt with a JSON output format achieved competitive worst-case accuracy across all model-by-task cells.
Based on these findings, we adopted a single default task prompt for all main experiments to ensure fair comparison.

\begin{figure}
  \centering
  \includegraphics[width=.97\textwidth]{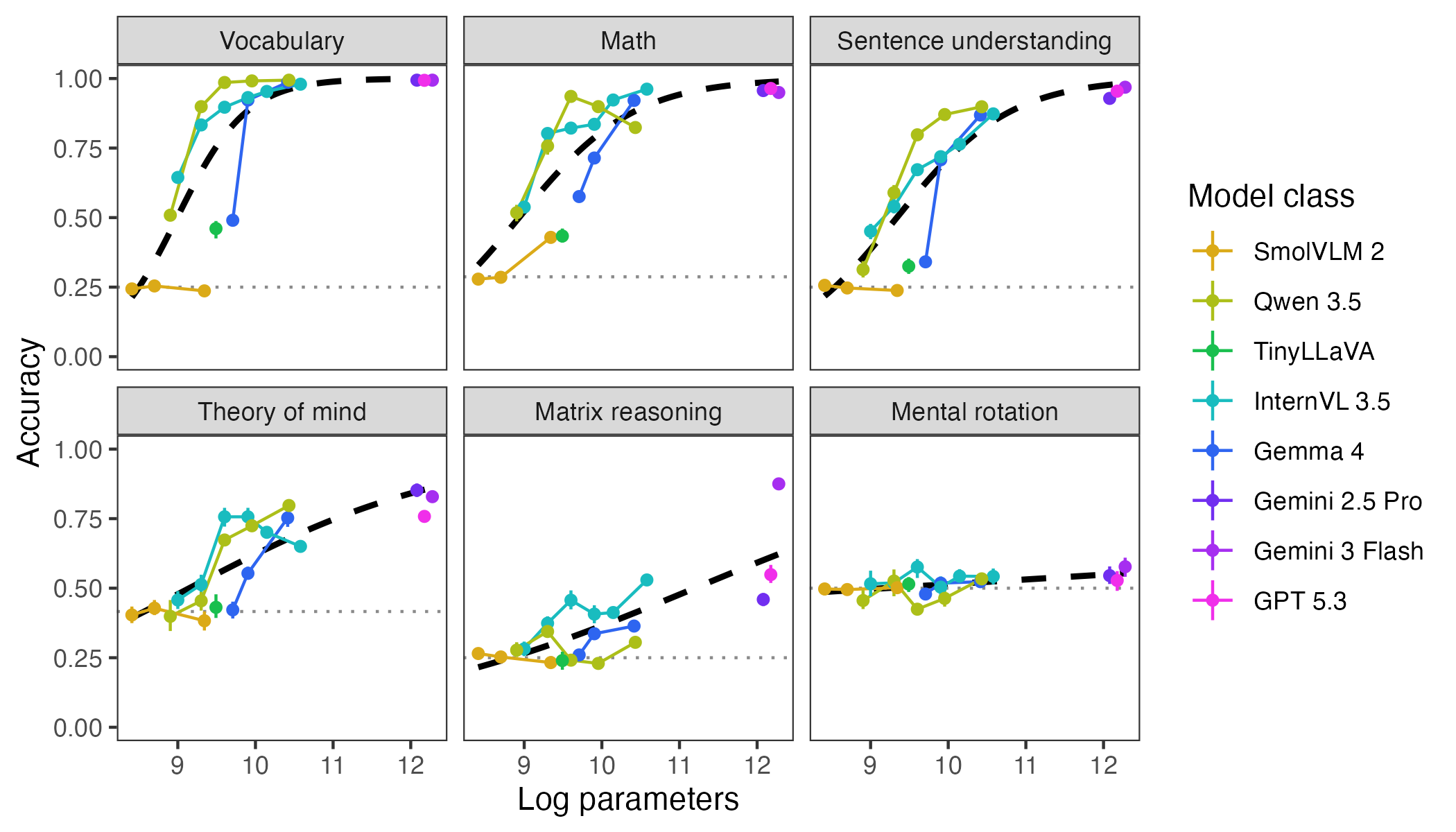}
  \caption{VLM accuracies plotted by log$_{10}$ parameters (estimated for commercial models). Error bars indicate bootstrapped 95\% confidence intervals. Colors denote model families. Dotted lines indicate chance levels, which vary between tasks due to varying numbers of response options. Dashed lines show best fitting logistic regressions. }
  \label{fig:accuracies}
\end{figure}

\section{Results}
\label{sec:results}

We first examine overall model accuracies on LEVANTE-bench in English (§\ref{sec:accuracy}). Next, we compare cross-task (§\ref{sec:cross-task}) and within-task (§\ref{sec:within-task}) alignment in accuracies as well as alignment in response distributions (§\ref{sec:error}). Finally, we report cross-linguistic results for German and Spanish (§\ref{sec:xling}).

\subsection{Model accuracy}
\label{sec:accuracy}

Figure~\ref{fig:graph_abs}B shows the overall performance of models and humans across all six tasks in LEVANTE-bench.
Figure~\ref{fig:accuracies} shows model performance on each task as a function of model class and the number of parameters of the model.
Recapitulating prior work showing systematic relationships between model size and performance \cite{ruanObservationalScalingLaws2024}, we found that larger models were more accurate across tasks, with performance following an approximately logistic form; this trend was also broadly observed within model classes.
Models performed relatively well in language tasks (vocabulary and sentence understanding) and in the math task, but poorly in spatial and relational reasoning tasks (mental rotation and matrix reasoning).
In particular, all models were approximately at chance for mental rotation, and even commercial frontier models did not reach high levels of performance (cf. \citep{stogiannidis2025mindgapbenchmarkingspatial}).

\begin{figure}
  \centering
  \includegraphics[width=\textwidth]{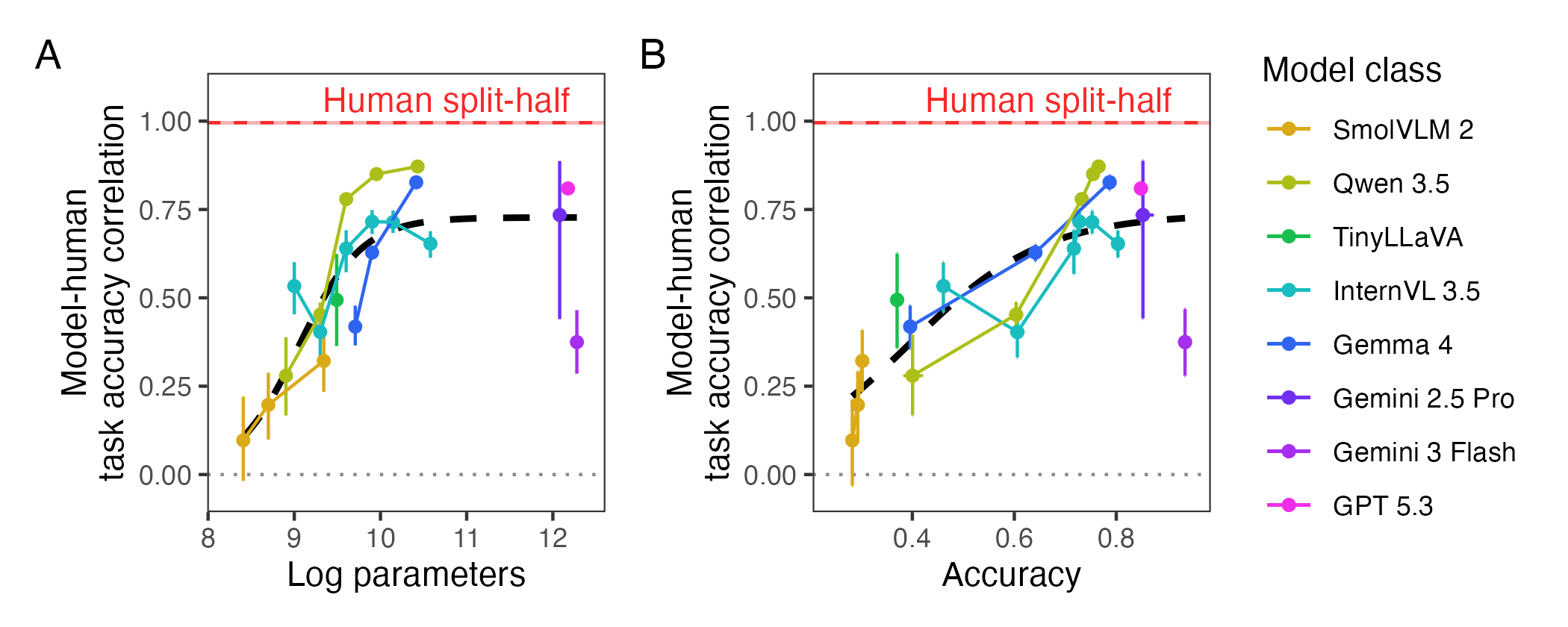}
  \caption{Correlation between model and human task accuracies plotted by (A) log$_{10}$ parameters and (B) overall accuracy. Error bars indicate bootstrapped 95\% confidence intervals. Dotted line indicates zero. Dashed black lines show best fitting sigmoid regressions. Dashed red line and shaded region indicate human split-half correlation and bootstrapped 95\% confidence interval.}
  \label{fig:tasks}
\end{figure}

\subsection{Cross-task alignment}
\label{sec:cross-task}

Across our six tasks, humans and models were positively correlated with respect to task difficulty. In fact, larger and better-performing models tended to be more correlated with humans (Figure~\ref{fig:tasks}), although even commercial models remained significantly below human split-half correlations, estimated by calculating the task accuracy correlation between splits for 1000 bootstrapped random splits of human participants. This gap appears to be driven by models' high performance on vocabulary and very low performance on mental rotation.

\subsection{Within-task alignment}
\label{sec:within-task}

\begin{figure}
  \centering
  \includegraphics[width=\textwidth]{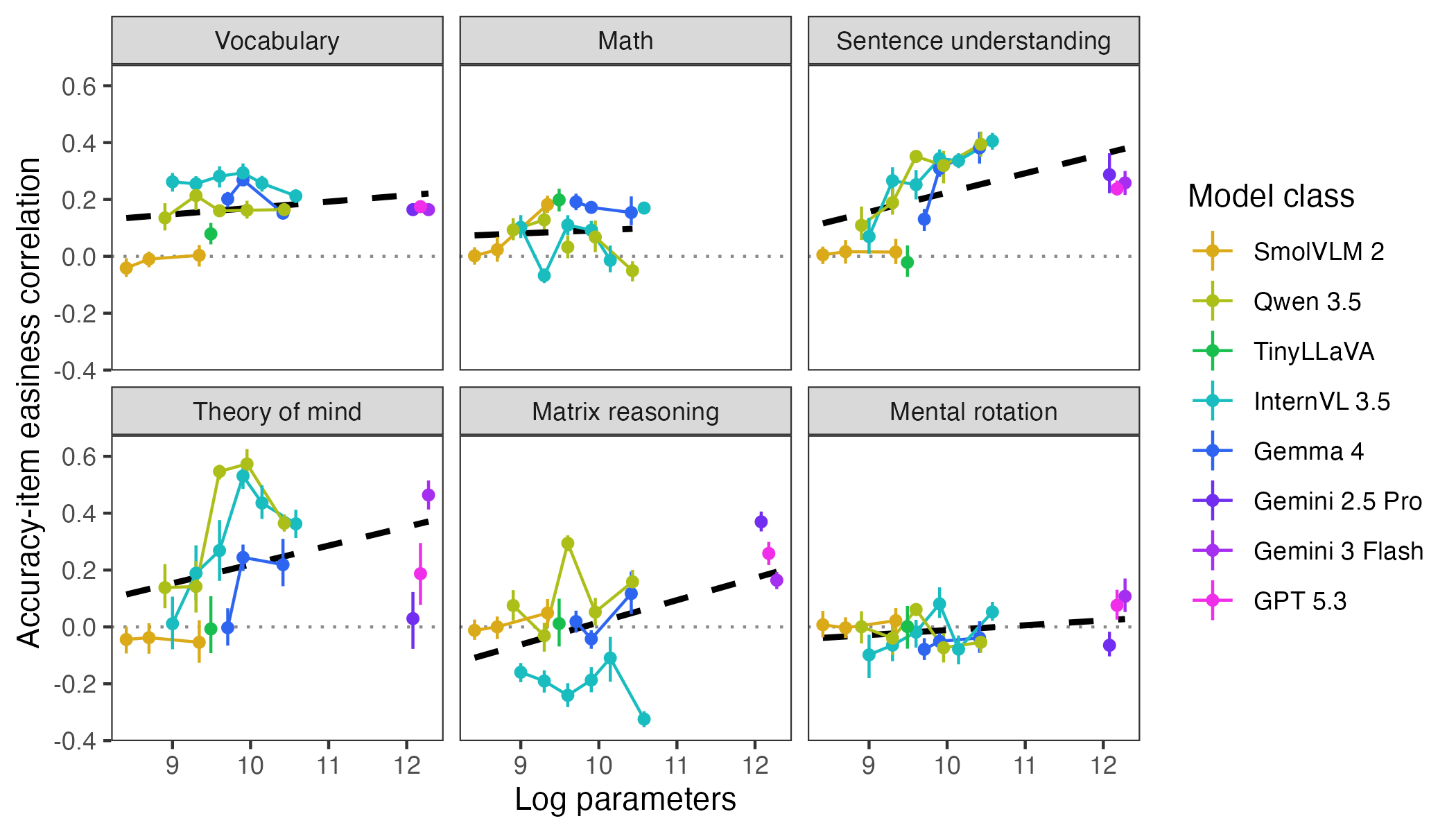}
  \caption{Correlation between model accuracy and item easiness estimated from human performance. Error bars indicate bootstrapped 95\% confidence intervals. Dotted lines indicate zero. Dashed lines show best fitting linear regressions. Correlations could not be estimated for the math task for GPT 5.3 and Gemini 2.5 Pro, as they had 100\% accuracy in retained items and thus no variance across items.}
  \label{fig:difficulties}
\end{figure}

Next, we examined model--human alignment on the relative item difficulties within a task.
Correlations were mostly positive but small across our models (Figure~\ref{fig:difficulties}); larger models showed slightly higher correlations with humans. 
% These results diverge notably from the task difficulty results: within-task alignment was more-highly correlated than cross-task alignment. 
% One possible explanation for this divergence relies on the observation that task-level difficulties may reflect the overall cognitive demands of the task domain (e.g., vocabulary recognition versus matrix reasoning), such that models and humans have fundamentally different learning trajectories across domains. 
% In contrast, item-level difficulties may reflect lower-level variation in perceptual, linguistic, or reasoning complexities within a single domain, affecting models and humans in a more similar manner.
Additional analysis on item subtypes (Appendix~\ref{sec:trial-type}) suggests that models and humans diverge on what kinds of trials are easy -- for example, humans are relatively better than models on addition but worse at fractions, and found 2D mental rotation easier while models found 3D easier; these differences resulted in relatively modest within-task alignment overall.

\subsection{Trial-level error alignment}
\label{sec:error}

Next, we investigated the alignment of response distributions between models and humans at the trial level, providing a finer-grained comparison on whether their error patterns are similar. We binned participants by ability (with $\theta$ bin size 1), and calculated the response distribution within each bin, excluding trials that were administered fewer than 10 times within a bin. We then calculated the Kullback--Leibler divergence ($D_\textup{\small KL}$) between model response patterns and human response patterns.

\begin{figure}
  \centering
  \includegraphics[width=\textwidth]{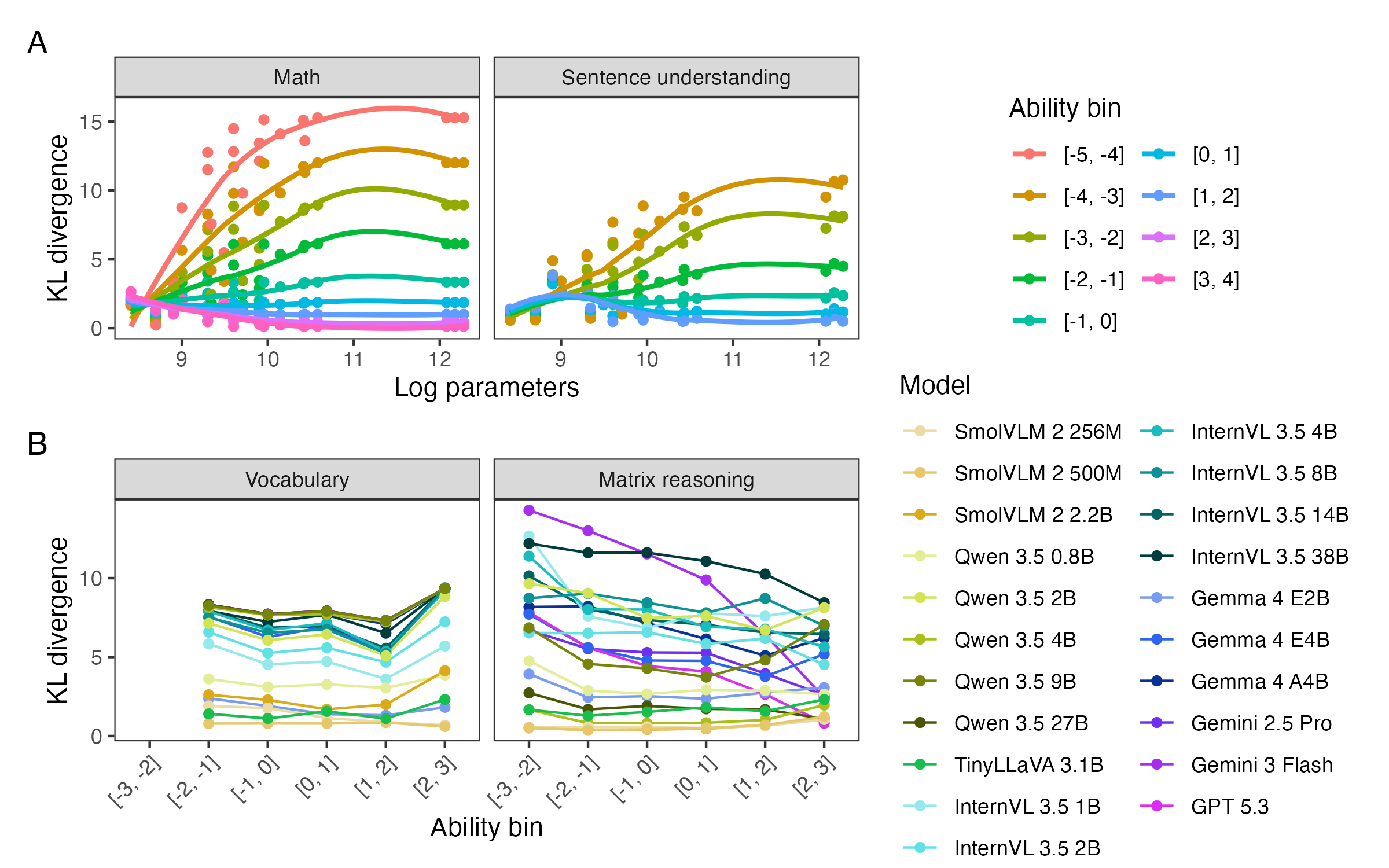}
  \caption{Tasks showed markedly different patterns of trial-level alignment. $D_\textup{\small KL}$ between model and human response distributions plotted by (A) log$_{10}$ number of parameters (for math and sentence understanding) and (B) human ability bins (for vocabulary and matrix reasoning). Lower values indicate greater model--human alignment.}
  \label{fig:distrib}
\end{figure}

Patterns of model--human divergence were highly heterogeneous between tasks (Figure~\ref{fig:distrib}), showing three patterns. First, math, sentence understanding, and theory of mind (not shown) showed a size effect: Larger models better matched higher-ability humans, while the smallest models matched lower-ability (low $\theta$) humans (who were often younger). This pattern suggests that smaller models, while lower accuracy, can still match lower-ability children's errors (replicating findings in \citep{tan2025devbench}). Second, for vocabulary and matrix reasoning, error pattern alignment was more model-specific, and $D_\textup{\small KL}$ tended to be similar across ability bins (with a few notable exceptions, including the largest models for matrix reasoning).
Third, mental rotation (not shown) had relatively low and constant $D_\textup{\small KL}$ values, perhaps due to overall poor differentiation across models on this task.
See Appendix~\ref{sec:d_kl} for full $D_\textup{\small KL}$ results.

\subsection{Cross-linguistic results}
\label{sec:xling}

\begin{figure}
    \centering
    \includegraphics[width=\linewidth]{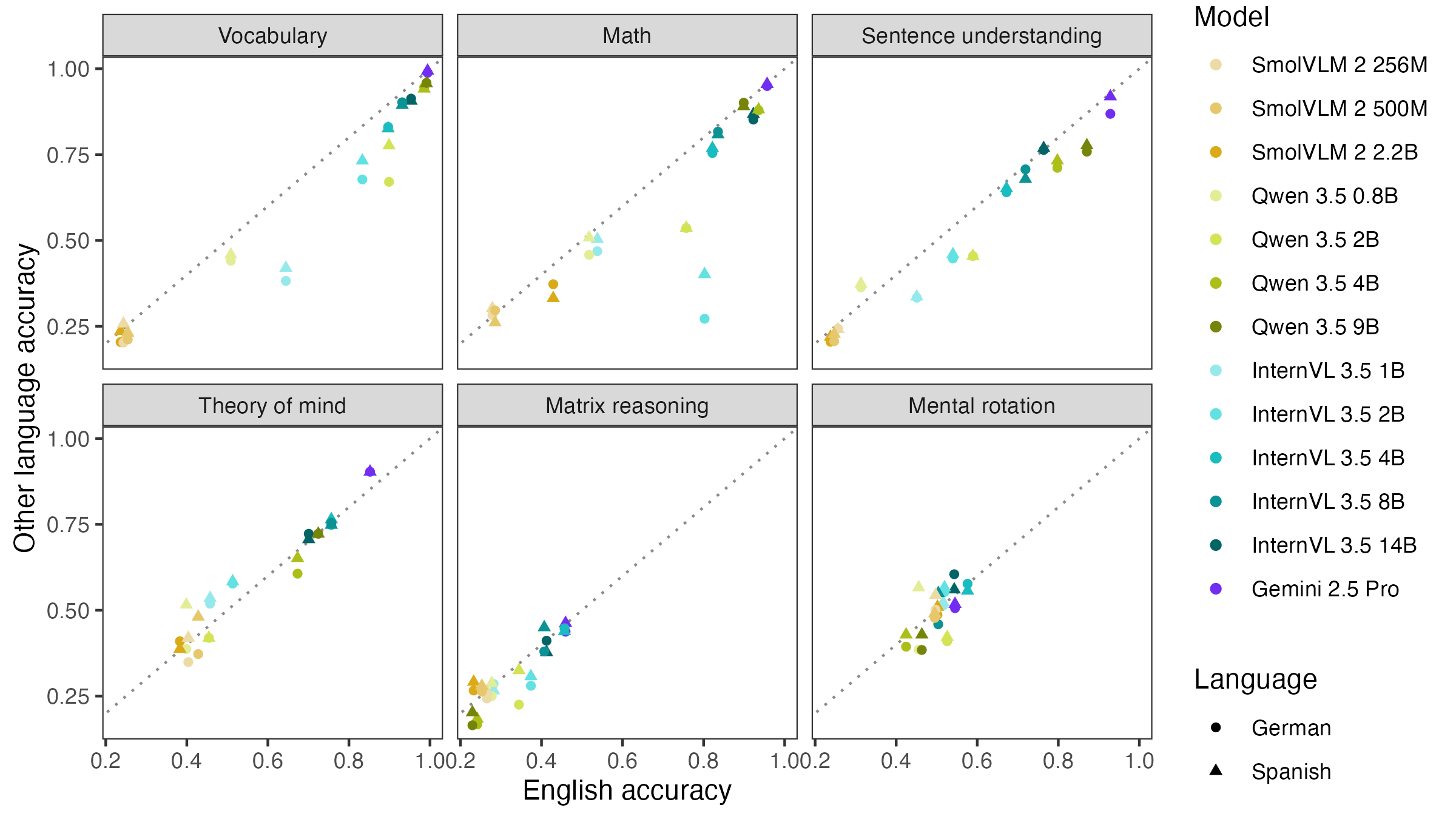}
    \caption{Comparison of model accuracies on English versus German/Spanish versions of our tasks.}
    \label{fig:xling}
\end{figure}

We additionally tested a subset of models using LEVANTE German and Spanish item translations.\footnote{Some models were run only once in German and Spanish, rather than 10 times as in English.} We used the professionally translated item prompts from the LEVANTE framework, and machine translation for the general task prompts. Models show relatively high correlation between English and German/Spanish accuracies across tasks (Figure~\ref{fig:xling}). However, many models showed lower vocabulary scores and some model sizes showed lower performance on the math task. Overall, however, these results demonstrate models' robustness across languages and suggest an opportunity for investigating cognitively plausible multilingual training regimes.\footnote{By design, the LEVANTE framework does not support human cross-site comparisons of cognitive ability due to many differences in sampling and administration across sites \citep{frank2025learning}. However, models which claim to be multilingual should have equitable access and performance across languages, permitting direct comparison on LEVANTE-bench.}

\section{Discussion}

Developing VLMs as models of human learning will require benchmarks to assess their alignment with children's behavior. These benchmarks must span across cognitive abilities and ages, and should ideally assess children across linguistic and cultural groups. LEVANTE-bench takes a step towards accomplishing these goals by leveraging a large, cross-national collaborative open science project to use validated tasks and a large dataset for human--model comparison. Critically, rather than examining only absolute task performance, LEVANTE-bench affords multi-scale comparison at the level of tasks, items, and error distributions. We found that current VLMs show moderate task-level alignment that nonetheless falls short of human–human reliability; modest item-level alignment in larger models; and heterogeneous trial-level alignment that depended on both task and model size. Thus, current models show only partial alignment with human cognitive development.

Some differences between models and children that we observed plausibly stem from the well-documented weaknesses of current VLMs in spatial cognition (e.g., \citep{stogiannidis2025mindgapbenchmarkingspatial,li2025coreknowledgedeficitsmultimodal}). For example, the failures of models in mental rotation tasks are not intrinsic to all vision models, as special-purpose networks can be highly successful \citep{mason2026largevisionmodelssolve}. Similarly, while multiplication and fraction trials in the math task were hard for humans, counting visual arrays was the most challenging for models \citep{qharabagh2026lvlmcountenhancingcountingability} (see Appendix~\ref{sec:trial-type}). Thus, developing better cognitive models will likely require vision backbones with stronger cognitive abilities \citep{aw2026zeroshotworldmodelsdevelopmentally,garrido2025intuitive}. 

\subsection{Limitations and future directions}

Because of the diversity of tasks, wide range of ages, and cross-linguistic diversity, LEVANTE-bench constitutes a substantial advance over previous developmental benchmarks. Nevertheless, it still has substantial limitations. The current dataset only includes data from six tasks and three languages. Work in progress seeks to increase the set of LEVANTE languages and to include tasks that span other key cognitive domains. Further, included tasks have a relatively small number of items (ranging from dozens in theory of mind to several hundred in math); this item set should be expanded to increase benchmark precision. Because LEVANTE is an open-source project, items are available on GitHub; despite its non-commercial license, it may still be included in the training data of some recent models, leading to possible contamination. Because our results do not center on absolute performance comparisons, we do not view this as a critical issue. 

Scaling LEVANTE-bench can be challenging as analyses of trial-level alignment require repeated sampling across models. While we generated at least 10 responses for all models, more precise distributional estimates require substantial computational resources relative to conventional single-pass benchmarks. Finally, we made a number of procedural decisions in preparing benchmark materials (including prompt content, image size, and output parsing). We attempted to make choices that limited worst-case behavior by smaller models (minimizing ``task demands'' per \citep{hu2024auxiliary}), but it is possible that specific optimizations could improve individual model families' performance. 

Finally, our eventual goal is to use our benchmark to help answer questions about human learning. For example, human alignment scores could be used to measure how VLMs' representations change with training or to compare VLMs with cognitively-relevant differences in architecture or training data. Achieving this goal depends on access to models that can be analyzed in this way, however. We did not have access to generative VLMs trained on data from children (cf. \citep{vong2024,wang2025bbabyvlm}), limiting our ability to make direct comparisons between humans and models trained on human data. Further, to our knowledge, no prominent VLM families have released checkpoint data, which would allow us to analyze alignment across training within a single model (as was done in \citep{tan2025devbench}).

\subsection{Conclusions}

Language models promise to provide an important tool for understanding human learning and cognition \citep{frank2025cognitive, connell2024can, mcgrath2024can}. Fulfilling this promise, however, will require models that learn from human-scale input \citep{frank2023bridging, vong2024, warstadt2023call} and that provide a good match to human learning trajectories from early childhood to adulthood. Benchmarks play a critical role in this process by furnishing the field with measurement instruments to better assess progress in model development. In particular, LEVANTE-bench allows for model--human comparisons at multiple scales -- cross-task, within-task, and trial-level error alignments -- using tasks that have been psychometrically validated in humans and have parallel versions in multiple languages. We hope that the current work will thus provide an improved measure with which to assess some of our progress towards models of the human mind. 

\begin{ack}
This work, including development and data collection for the LEVANTE core tasks, was supported by the Jacobs Foundation. Gemini credits were provided through a gift from Google, Inc. Some of the computing for this project was performed on the Marlowe cluster. We would like to thank Stanford University and Stanford Research Computing for providing computational resources and support that contributed to these research results, and Mika Braginsky for help with data processing.

\paragraph{Author contributions.} \emph{Conceptualization}: AWMT, DC, TLB, LBS, MCF; \emph{Data curation}: AWMT, MCF; \emph{Formal analysis}: AWMT, TLB, SY; \emph{Funding acquisition}: MCF; \emph{Investigation}: AWMT, DC, TLB, LBS, MCF; \emph{Methodology}: AWMT, DC, TLB, LBS, MCF; \emph{Resources}: DC, MCF; \emph{Software}: AWMT, DC, TLB, LBS, SY, MCF; \emph{Supervision}: MCF; \emph{Visualization}: AWMT, TLB, SY; \emph{Writing -- original draft}: AWMT, MCF; \emph{Writing -- review \& editing}: AWMT, DC, TLB, LBS, SY, MCF.

\paragraph{Artificial intelligence disclosure.} \emph{Artificial intelligence tools}: ChatGPT, Claude Code, Claude Cowork, Cursor, Gemini 2.5 Pro, and Perplexity; \emph{Data collection methods}: Gemini 2.5 Pro was used to machine translate the task prompts into German and Spanish; \emph{Execution}: ChatGPT, Claude Code, and Cursor were used to assist in code generation for running experiments; \emph{Writing -- original draft}: Cursor was used to help with writing Appendix~\ref{sec:prompt_sensitivity}; \emph{Writing -- review \& editing}: Claude Cowork and Perplexity were used to review the manuscript for coherence and clarity.

\end{ack}

% \section*{References}

% {
% \small
\bibliography{levante_bench.bib}

% }

%%%%%%%%%%%%%%%%%%%%%%%%%%%%%%%%%%%%%%%%%%%%%%%%%%%%%%%%%%%%
\newpage
\appendix

% \section{Technical appendices and supplementary material}

% \subsection{Multi-image prompt alternative for vocabulary}

% Given the poor (near-chance) performance of our smallest models on the vocab task, we speculated that feeding them four images exceeded their effective memory. So we built a collection of 4 images in quadrants (exactly what is shown to a child) as alternative. However, this approach didn't result in any improvement in results. We have provided that version of our vocab items in an additional\_images folder in our levante\_bench Google bucket for researchers who want to do further experiments.

% \begin{taniablock}
\section{Prompt sensitivity analysis}
\label{sec:prompt_sensitivity}

We conducted a systematic exploration of prompt design across all six tasks and multiple model families before selecting the default prompt for the main experiments.
The goal was to verify that our results reflect model capabilities rather than artifacts of a particular prompt phrasing.
All notebooks and raw output are available in our supplementary materials.

\paragraph{Methodology.}
For each task we defined a set of prompt \emph{phases} that could be applied independently or in combination:
\textbf{Phase~0}~(baseline): the task stimulus followed by a JSON output instruction;
\textbf{Phase~1}~(structured layout): reformatted multiline text with labeled option blocks;
\textbf{Phase~2}~(enhanced parsing): parser-side improvements only (no prompt change);
\textbf{Phase~3}~(task system prompt): a task-specific system message (e.g., ``You are a visual vocabulary expert'');
\textbf{Phase~4}~(task-specific hints): domain knowledge such as mirror/chirality hints for mental rotation or distractor-awareness cues for vocabulary;
\textbf{Phase~5}~(chain-of-thought): step-by-step reasoning instructions, sometimes with increased token budgets (512--1024 tokens).
All phases were tested individually and in combinations up to the full stack.

\paragraph{Cross-model sweep.}
A dedicated robustness sweep tested 5~models (Qwen~0.8B, InternVL~1B, SmolVLM2~2.2B, InternVL~4B, and Gemma3~4B) across all 6~tasks using 4~task-framing variants $\times$ 2~output-format variants (bare letter vs.\ JSON), for a total of 240~model--task--prompt cells.
The overall mean accuracy was 46.9\% (bare) vs.\ 45.7\% (JSON), and mean parse rates were 98.1\% vs.\ 96.8\%, indicating that the output format choice has a small effect.
A maximin analysis--selecting the prompt that maximizes the worst-case accuracy across models--favored the \emph{minimal framing with JSON output} (TF1$\times$OF2) on 4 of 6 tasks.

\paragraph{Per-task findings.}
Table~\ref{tab:prompt_sensitivity} summarizes the key results.

\begin{table}[h]
  \centering
  \caption{Prompt sensitivity by task: accuracy range observed across prompt strategies for representative models. $\Delta_{\max}$ is the spread between the best and worst prompt configuration.}
  \label{tab:prompt_sensitivity}
  \small
  \begin{tabular}{llcccc}
    \toprule
    Task & Model & Baseline & Best prompt & Worst prompt & $\Delta_{\max}$ \\
    \midrule
    Math & Qwen 0.8B & 29.6\% & 57.0\% (CoT) & 27.9\% (few-shot) & 29.1 pp \\
    Vocabulary & Qwen 0.8B & 71.8\% & 84.1\% (full stack) & 48.2\% (CoT only) & 35.9 pp \\
    Sentence und. & Qwen 0.8B & 36.4\% & 43.4\% (struct.) & 26.3\% (CoT) & 17.1 pp \\
    Sentence und. & Qwen 2B & 51.5\% & 64.6\% (struct.+desc.) & 43.4\% (expert) & 21.2 pp \\
    Theory of mind & InternVL 4B & 45.9\% & 59.5\% (system+anchor) & 40.5\% (hints) & 19.0 pp \\
    Matrix reasoning & InternVL 2B & 38.0\% & 41.8\% (struct.+hints) & 17.7\% (expert) & 24.1 pp \\
    Matrix reasoning & InternVL 8B & 44.3\% & 57.0\% (struct.+expert) & 44.3\% (baseline) & 12.7 pp \\
    Mental rotation & Qwen 0.8B & 59\% & 60\% (struct.) & 59\% (baseline) & 1 pp \\
    Mental rotation & Qwen2.5 3B & 48.2\% & 62.7\% (SC k=5) & 0\% (perm. debias) & 62.7 pp \\
    \bottomrule
  \end{tabular}
\end{table}

Several patterns emerged consistently:

\begin{enumerate}
  \item \textbf{CoT hurts small models.} For sub-2B models, chain-of-thought instructions reduced both accuracy and parse rate on vocabulary ($-$23.5~pp), sentence understanding ($-$10.1~pp), and matrix reasoning ($-$7.6~pp). Extended reasoning consumed the token budget without producing a parseable answer.

  \item \textbf{Prompt gains are model-specific.} The expert system prompt that boosted InternVL~8B on matrix reasoning by +12.7~pp \emph{decreased} InternVL~2B accuracy by $-$20.3~pp on the same task. Similarly, the describe-first strategy that helped Qwen~2B on sentence understanding (+13.1~pp) was not effective for the 0.8B variant.

  \item \textbf{Mental rotation resists prompting.} Across Qwen~0.8B, InternVL~2B, InternVL~8B, and three spatial fine-tuned models (SpaceThinker, SpaceOm, SpatialThinker), no prompt strategy reliably exceeded chance after controlling for position bias via answer-permutation debiasing. The apparent best result (62.7\% via self-consistency on Qwen2.5-VL-3B) was not significant under a bias-aware null model ($p \approx 0.183$).

  \item \textbf{Spatial fine-tuning does not help.} Three models fine-tuned for spatial reasoning (SpaceThinker, SpaceOm, SpatialThinker-Oxford) all scored 59.0\% with the baseline elimination prompt---identical to the position-biased ceiling---and dropped to 38.6--48.2\% with their recommended paper prompts, which elicit longer reasoning chains.

  \item \textbf{Stacking phases has diminishing or negative returns.} Full-stack combinations often underperformed the best single phase. In sentence understanding (0.8B), combining all phases yielded 35.4\% vs.\ 43.4\% for the structural stack alone. In vocabulary, the full stack was the exception that improved over individual phases (+12.4~pp), driven by synergistic interactions between structure and distractor awareness.
\end{enumerate}

\paragraph{Justification for the default prompt.}
Given these results, we opted for the minimal JSON-based prompt as the default for main experiments. This configuration (i)~achieves competitive accuracy across all model families in maximin analysis, (ii)~maintains parse rates above 95\% for all models except the smallest, and (iii)~enables a fair, model-agnostic comparison without introducing prompt-induced variance that could confound cross-model and cross-task analyses.
% \end{taniablock}

\section{Additional results}

\subsection{Full KL divergence results}
\label{sec:d_kl}

\begin{figure}
    \centering
    \includegraphics[width=\linewidth]{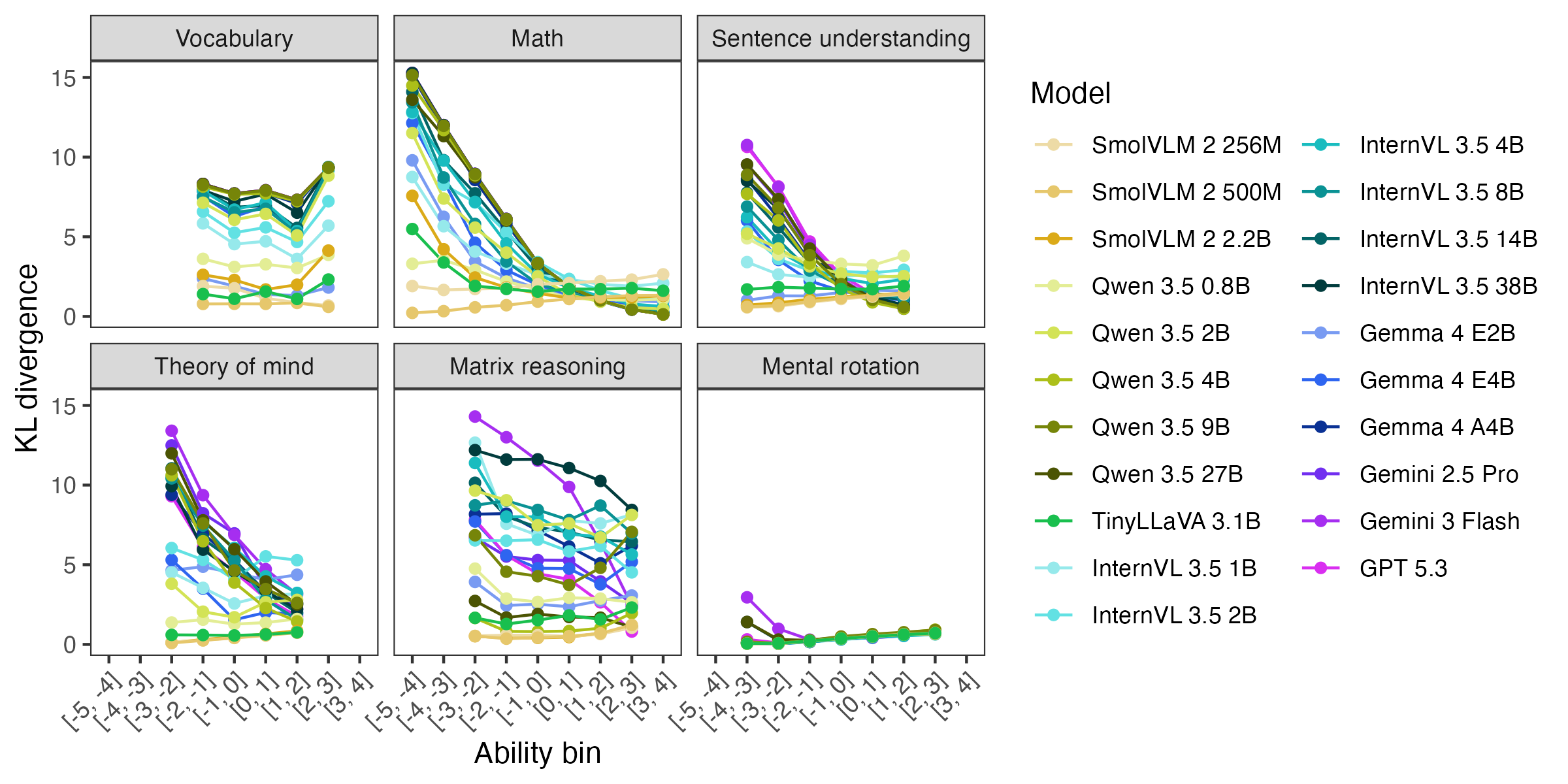}
    \caption{$D_\textup{\small KL}$ between model and human response distributions plotted by log$_{10}$ number of parameters for all tasks. Lower values indicate greater model--human alignment.}
    \label{fig:distrib_params}
\end{figure}

\begin{figure}
    \centering
    \includegraphics[width=\linewidth]{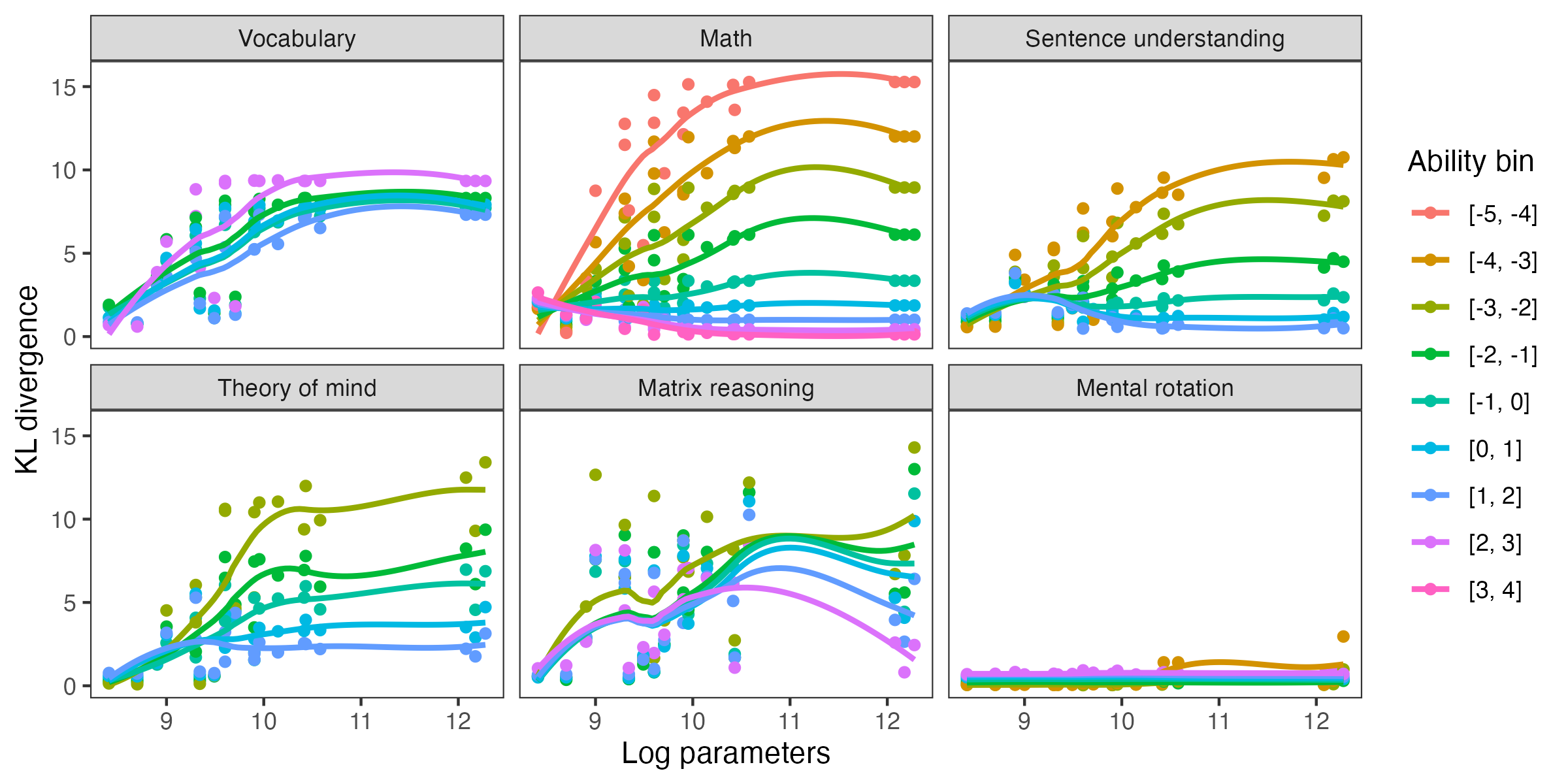}
    \caption{$D_\textup{\small KL}$ between model and human response distributions plotted by human ability bins for all tasks. Lower values indicate greater model--human alignment.}
    \label{fig:distrib_ability}
\end{figure}

Figures~\ref{fig:distrib_params} and~\ref{fig:distrib_ability} show the full $D_\textup{\small KL}$ distributions for all models across all ability bins for all tasks. Math, sentence understanding, and theory of mind show similar trends, with larger models being more similar to higher ability humans and vice versa. Vocabulary and matrix reasoning show $D_\textup{\small KL}$ distributions that depend more on the specific model, while mental rotation shows collapse across models with very limited differentiation.

\subsection{Item subtype analysis}
\label{sec:trial-type}

\begin{figure}
    \centering
    \includegraphics[width=\linewidth]{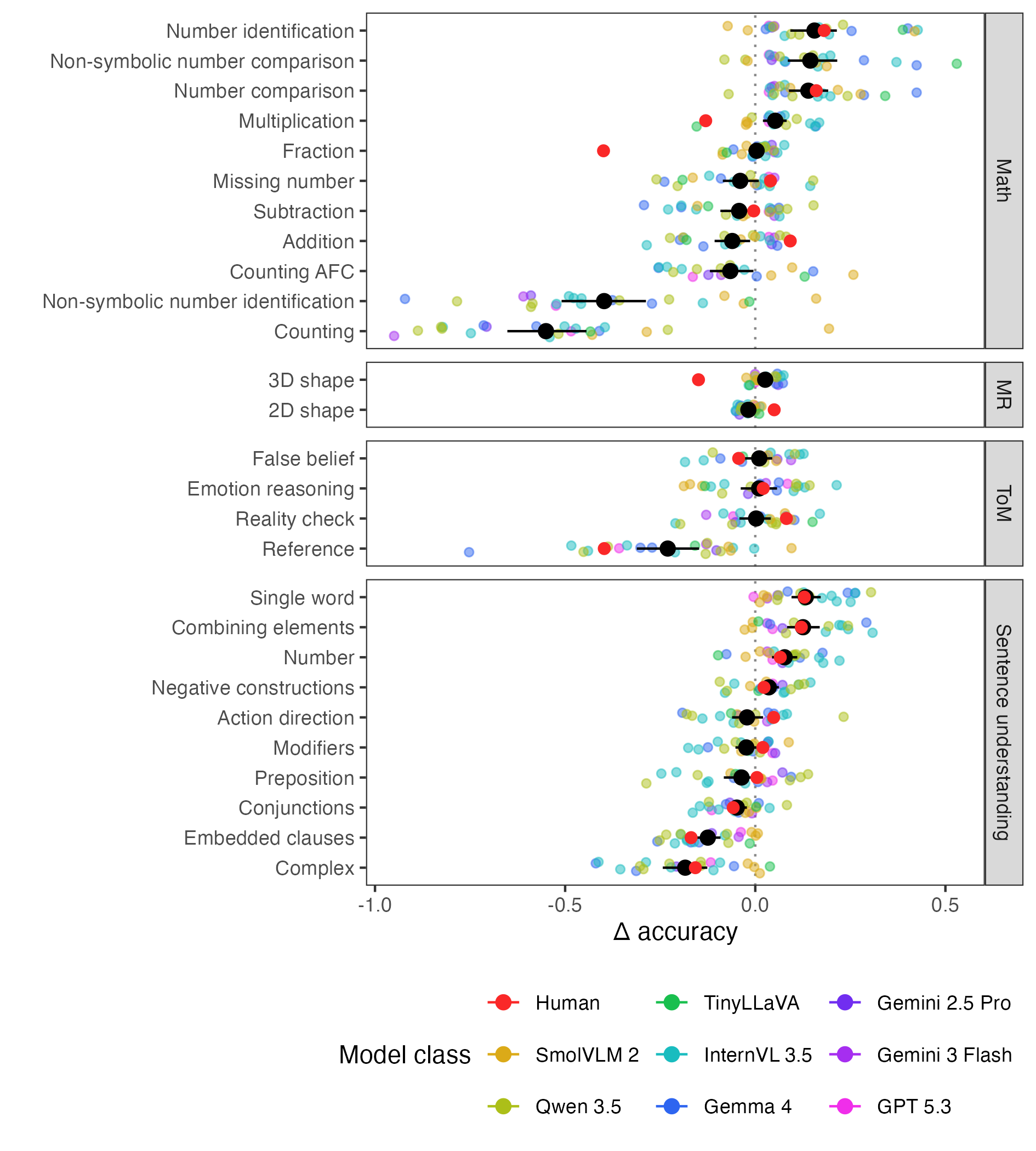}
    \caption{Accuracy deviation from overall task accuracy. Black dots and error bars indicate all-model means and bootstrapped 95\% confidence intervals. Dotted line indicates zero. Human accuracy estimates are not yet available for some item subtypes. MR: mental rotation; ToM: theory of mind.}
    \label{fig:placeholder}
\end{figure}

To understand the distribution of model accuracies across items, we further examined model performance for tasks with item subtypes. We calculated each model's accuracy for each item subtype (for tasks with subtypes), then calculated the difference between subtype accuracy and overall task accuracy. 

This analysis revealed areas of divergence between models and humans. For the math task, humans were comparatively better at addition and missing number tasks relative to models, whereas they were comparatively much worse at multiplication and fraction questions. For mental rotation, humans found 2D shapes easier than 3D shapes, whereas the reverse was true for models. For theory of mind, humans showed a relative ordering between reality check, emotion reasoning, and false believe questions, whereas models were broadly similar on these three subtypes. Interestingly, sentence understanding showed relatively similar deviations in subtype accuracy between models and humans. 
These disparities suggest that items may function differentially between models and humans, especially in reasoning domains.

%%%%%%%%%%%%%%%%%%%%%%%%%%%%%%%%%%%%%%%%%%%%%%%%%%%%%%%%%%%%

% \newpage
% \input{checklist.tex}

\end{document}